\documentclass[letterpaper]{article} 
\usepackage{aaai23}  
\usepackage{times}  
\usepackage{helvet}  
\usepackage{courier}  
\usepackage[hyphens]{url}  
\usepackage{graphicx} 
\usepackage{natbib}  
\usepackage{caption} 
\usepackage{enumitem}
\usepackage{algorithm}
\usepackage{algorithmic}
\usepackage{graphicx}
\usepackage{subfigure}
\usepackage{booktabs} 
\usepackage{amsmath}
\usepackage{amssymb}
\usepackage{booktabs}
\usepackage{multirow}
\usepackage{xcolor}
\usepackage{caption}
\usepackage{mathrsfs}  
\definecolor{applegreen}{rgb}{0.0, 0.5, 0.0}
\usepackage{colortbl}
\usepackage{mathtools}
\DeclarePairedDelimiter{\ceil}{\lceil}{\rceil}
\usepackage[normalem]{ulem}

\newtheorem{definition}{Definition} 

\newcommand{\YGL}[1]{{{\textcolor{blue}{\textbf{YGL:}}}{\textcolor{red}{\textbf{#1}}}}}
\newcommand{\YC}[1]{{{\textcolor{black}{\textbf{YC:}}}{\textcolor{blue}{\textbf{#1}}}}}

\usepackage{newfloat}
\usepackage{listings}
\DeclareCaptionStyle{ruled}{labelfont=normalfont,labelsep=colon,strut=off} 
\floatstyle{ruled}
\newfloat{listing}{tb}{lst}{}
\floatname{listing}{Listing}
\title{Topological Pooling on Graphs}
\author{
    Yuzhou Chen,$^1$
    Yulia R. Gel$^{2,3}$
}
\affiliations{
    \textsuperscript{\rm 1}Department of Computer and Information Sciences, Temple University\\
    \textsuperscript{\rm 2}Department of Mathematical Sciences, University of Texas at Dallas\\
    \textsuperscript{\rm 3}National Science Foundation\\
    yuzhou.chen@temple.edu, ygl@utdallas.edu\\
}

\usepackage{bibentry}

\begin{document}
\maketitle

\begin{abstract}
Graph neural networks (GNNs) have demonstrated a significant success in various graph learning tasks, from graph classification to anomaly detection. There recently has emerged a number of approaches adopting a graph pooling operation within GNNs, with a goal to preserve graph attributive and structural features during the graph representation learning. However, most existing graph pooling operations suffer from the limitations of relying on node-wise neighbor weighting and embedding, which leads to insufficient encoding of rich topological structures and node attributes exhibited by real-world networks. By invoking the machinery of persistent homology and the concept of landmarks, we propose a novel topological pooling layer and witness complex-based topological embedding mechanism that allow us to systematically integrate hidden topological information at both local and global levels. Specifically, we design new learnable local and global topological representations Wit-TopoPool which allow us to simultaneously extract rich discriminative topological information from graphs. Experiments on 11 diverse benchmark datasets against 18 baseline models in conjunction with graph classification tasks indicate that Wit-TopoPool significantly outperforms all competitors across all datasets.

\end{abstract}

\section{Introduction}
Graph neural networks (GNNs) have emerged as a powerful machinery for various graph learning tasks, 
such as link prediction, node and graph classification~\cite{zhou2020graph, xia2021graph}. 
In case of graph classification tasks, both graph-level and local-level 
representation learning play a critical role for GNNs success. Since pooling operators have shown an important role in image recognition and 3D shape analysis
~\cite{boureau2010theoretical,shen2018mining}, it appeared natural to expand the idea of pooling to graphs~\cite{yu2016multi,defferrard2016convolutional}. The earlier pooling techniques within GNN architectures achieved promising results but failed to capture graph structural information and to learn hidden node feature representations. These limitations have stimulated development of new families of graph pooling which simultaneously   
account for both the structural properties of graphs and node feature representations. Some of the most recent results on graph pooling in this direction include integration of (self)-attention mechanisms~\cite{huang2019attpool,lee2019self}, advanced clustering approaches~\cite{bianchi2020spectral,wang2020haar,bodnar2021deep}, and hierarchical graph representation learning~\cite{yang2021hierarchical}.   
Nevertheless, these existing pooling techniques still remain limited in their ability to simultaneously extract and systematically integrate intrinsic structural information on the graph, including its node feature representations, at both local and global levels. 

We address these limitations by introducing the concepts of shape, landmarks, and witnesses  to graph pooling. In particular, our key idea is based on the two interlinked notions from computational topology: persistent homology and witness complexes, neither of which has ever been considered in conjunction with graph pooling. First, we propose a local topological score which measures each node importance based on how valuable shape information of its local vicinity is and which then results in a topological pooling layer. Second, inspired by the notion of landmarks in computer vision, we learn the most intrinsic global topological properties of the graph, using only a subset of the most representative nodes (or landmarks).
The remaining nodes act as witnesses that govern which higher-order graph structures (or simplices) are to be included into the learning process. In computational topology, this approach is associated with a witness complex which enjoys competitive computational costs. The resulting new model Wit-TopoPool exhibits capabilities to learn rich discriminative topological characteristics of the graph as well as to extract essential information from node features. Significance of our contributions are the following:
\begin{itemize}[leftmargin=*]
\item We propose a novel topological perspective to graph pooling by introducing the concepts of persistence, landmarks, and witness complexes.

\item We develop a new model Wit-TopoPool which systematically and simultaneously integrates the most essential discriminative topological characteristics of the graph, including its node feature representations, at both local and global levels.

\item We validate Wit-TopoPool in conjunction with graph classification tasks versus 18 state-of-the-art competitors on 11 diverse benchmark datasets from chemistry, bioinformatics, and social sciences. 
Our experiments indicate that Wit-TopoPool delivers the most competitive performance under all considered scenarios.

\end{itemize}

\section{Related Work}

\paragraph{Graph Pooling and Graph Neural Networks}
In the last few years GNNs have proven to become the primary machinery for graph classification tasks~\cite{zhou2020graph, xia2021graph}.
Inspired by the success of GNN-based models for graph representation learning, there has appeared a number of approaches introducing a pooling mechanism into GNNs which addresses the limitations of traditional graph pooling architectures, i.e., a limited capability to capture the graph substructure information. For instance, DiffPool~\cite{ying2018hierarchical} develops a differential pooling operator that learns a soft assignment at each graph convolutional layer. The similar idea is utilized in EigenGCN~\cite{ma2019graph}, which introduces a pooling operator based on the graph Fourier transform. Top-$K$ pooling operations~\cite{cangea2018towards,gao2019graph} design a pooling method by using node features and local structural information to propagate only the top-$K$ nodes with the highest scores at each pooling step. Self-Attention Graph Pooling (SAGPool)~\cite{lee2019self} leverages self-attention mechanism based on GNN to learn the node scores and select the nodes by sorting their scores. Although these GNNs and graph pooling operations have achieved state-of-the-art performance in graph classification tasks, a common limitation among all aforementioned approaches is that they cannot accurately capture higher-order properties of graphs and incorporate this topological information into neural networks. Different from these existing methods, we propose a novel model Wit-TopoPool that not only adaptively captures local topological information in the hidden representations of nodes but yields a faster and scalable approximation of the simplicial representation of the global topological information.

\paragraph{Witness Complexes, Landmarks, and Topological Graph Learning}
Persistent homology (PH) is a suite of tools within topological data analysis (TDA) that has shown substantial promise in a broad range of domains, from bioinformatics to material science to social networks~\cite{otter2017roadmap,carlsson2020topological}. PH has been also successfully integrated as a fully trainable topological layer into various DL models, for such graph learning tasks as node and graph classification, 
link prediction 
and anomaly detection 
(see, e.g., overviews~\citet{carlsson2020topological,tauzin2021giotto}.
In most applications, PH is based on a Vietoris-Rips ($\mathcal{VR}$) complex which enjoys a number of important theoretical properties on approximation of the underlying topological space. However, $\mathcal{VR}$ does not scale efficiently to larger datasets and, in general, computational complexity remains one of the primary roadblocks on the way of the wider applicability of PH. A promising but virtually unexplored  alternative to $\mathcal{VR}$ is a witness complex~\cite{de2004topological} which recovers the topology of the underlying space using only a subset of points (or landmarks), thereby, substantially reducing computational complexity and potentially allowing us to focus only on the topological information delivered by the most representative points. Nevertheless, applications of witness complex in machine learning are yet nascent~\cite{schonenberger2020witness, poklukar2021geomca}. Here we harness strengths of witness complex and graph landmarks to accurately and efficiently learn topological knowledge representations of graphs.

\section{Methodology}
Let $\mathcal{G} = (\mathcal{V}, \mathcal{E}, \boldsymbol{X})$ be an attributed graph, where $\mathcal{V}$ is a set of nodes ($|\mathcal{V}|=N$), $\mathcal{E}$ is a set of edges, and $\boldsymbol{X} \in \mathbb{R}^{N \times F}$ is a node feature matrix (here $F$ is the dimension of node features). Let $d_{uv}$ be the distance on $\mathcal{G}$ defined as the shortest path between nodes $u$ and $v$, $u,v\in \mathcal{V}$, and $\boldsymbol{A} \in \mathbb{R}^{N \times N}$ be a symmetric adjacency matrix such that
$\boldsymbol{A}_{uv} = \omega_{uv}$ if nodes $u$ and $v$ are connected and 0, otherwise 
(here $\omega_{uv}$ is an edge weight and $\omega_{uv}\equiv 1$ for unweighted graphs).
Furthermore, $\boldsymbol{D}$ represents the degree matrix with $\boldsymbol{D}_{uu} = \sum_v \boldsymbol{A}_{uv}$, corresponding to $\boldsymbol{A}$. 
\begin{definition}[\footnotesize{$k$-hop Neighborhood based on Graph Structure}]
\label{def1}
An induced subgraph $\mathcal{G}^k_u=(\mathcal{V}^k_u, \mathcal{E}^k_u)\subseteq \mathcal{G}$ is called a $k$-hop neighborhood of node $u\in \mathcal{V}$ if for any $v\in \mathcal{V}^k_u$, $d_{uv}\leq k$. 
\end{definition}

\subsection{Preliminaries on Persistent Homology and Witness Complexes}
PH is a subfield in computational topology which allows us to retrieve evolution of the inherent shape patterns in the data along various user-selected geometric dimensions~\cite{edelsbrunner2000topological,zomorodian2005computing}. Broadly speaking, by ``shape" here we mean the 
properties of the observed object which are preserved under continuous transformations, e.g., stretching, bending, and twisting. (The data can be a graph, a point cloud in Euclidean space, or a sample of points from any metric space). Since one of the most popular PH techniques is to convert the point cloud to a distance graph, for generality we proceed with the further description of PH on graph-structured data. By using a multi-scale approach to shape description, PH enables to address the intrinsic limitations of classical homology and to extract the shape characteristics which play an essential role in a given learning task. In brief, the key idea is to choose some suitable 
scale parameters $\alpha$ and then to study changes in homology that occur to $\mathcal{G}$ which evolves with respect to $\alpha$. 
That is, we no longer study $\mathcal{G}$ as a single object but as a {\it filtration} $\mathcal{G}_{\alpha_1} \subseteq \ldots \subseteq \mathcal{G}_{\alpha_n}=\mathcal{G}$, induced by monotonic changes of $\alpha$. To make the process of pattern counting more systematic and efficient, we build an abstract simplicial complex $\mathscr{K}(\mathcal{G}_{\alpha_j})$ on each $\mathcal{G}_{\alpha_j}$, resulting in a filtration of complexes $\mathscr{K}(\mathcal{G}_{\alpha_1}) \subseteq \ldots \subseteq \mathscr{K}(\mathcal{G}_{\alpha_n})$.
For instance, we can select a scale parameter as 
a shortest (weighted) path between any two nodes; then
abstract simplicial complex $\mathscr{K}(\mathcal{G}_{\alpha_{*}})$ is generated by subgraphs $\mathcal{G}^{'}$ of bounded diameter $\alpha_{*}$ (that is, $(k-1)$-simplex in $\mathscr{K}(\mathcal{G}_{\alpha_{*}})$ is made up by subgraphs $\mathcal{G}^{'}$ of  $k$-nodes with  $diam(\mathcal{G}^{'})\leq \alpha_{*}$).
If $\mathcal{G}$ is an edge-weighted graph $(\mathcal{V}, \mathcal{E}, w)$, with the edge-weight function $w: \mathcal{E} \mapsto \mathbb{R}$, then 
for each $\alpha_j$ we can consider only induced subgraphs of $\mathcal{G}$ with maximal degree of $\alpha_j$, resulting in a degree sublevel set filtration.
(For the detailed discussion on graph filtrations see~\citet{hofer2020graph}.) 

Equipped with this construction, we trace data shape patterns such as independent components, holes, and cavities which appear and merge as scale $\alpha$ changes (i.e., for each topological feature $\rho$ we record the indices $b_{\rho}$ and $d_{\rho}$ of $\mathscr{K}(\mathcal{G}_{b_{\rho}})$ and $\mathscr{K}(\mathcal{G}_{d_{\rho}})$, where $\rho$ is first and last observed, respectively). We say that
a pair $(b_{\rho}, d_{\rho})$ represents the birth and death times of $\rho$, and $(d_{\rho} - b_{\rho})$ is its corresponding lifespan (or persistence). In general, topological features with longer lifespans are considered valuable, while features with shorter lifespans are often associated with topological noise. The extracted topological information over the filtration $\{\mathscr{K}_{\alpha_j}\}$ can be then summarized as a multiset in $\mathbb{R}^2$ called {\it persistence diagram (PD)}
$\mathcal{\text{Dg}}=\{(b_{\rho},d_{\rho}) \in \mathbb{R}^2: d_{\rho} > b_{\rho}\}\cup \Delta$ (here $\Delta= \{(t, t) | t \in \mathbb{R}\}$ is the diagonal set containing points counted with infinite multiplicity; including $\Delta$ allows us to compare different PDs based on the cost of the optimal matching between their points). 

Finally, there are multiple options to select an abstract simplicial complex $\mathscr{K}$~\cite{carlsson2021topological}. Due to its computational benefits, one of the most widely adopted choices is a Vietoris-Rips ($\mathcal{VR}$) complex.
However, the $\mathcal{VR}$ complex uses the entire observed data to describe the underlying topological space and, hence, does not efficiently scale to large datasets and noisy datasets. In contrast, a witness complex captures the shape structure of the data based only on a significantly smaller subset $\mathfrak{L}\subseteq \mathcal{V}$, called a set of {\it landmark} points. In turn, all other points in $\mathcal{V}$ are used as ``witnesses" that govern which simplices occur in the witness complex.
\begin{definition}[\footnotesize{Weak Witness Complex}]
\label{def2}
We call $w\in \mathcal{V}$ to be a {\it weak witness} for a simplex $\sigma=[v_0 v_1 \ldots v_l]$, where $v_i\in \mathcal{V}$ for $i=0,1,\ldots, l$ and nonnegative integer $l$, with respect to $\mathfrak{L}$ if and only if $d_{wv} \leq d_{wu}$ for all $v\in \sigma$ and $u \in  \mathfrak{L} \setminus \sigma$. The {\it weak witness complex} $\mathcal{W}(\mathfrak{L}, \mathcal{G})$ of the graph $\mathcal{G}$ with respect to $\mathfrak{L}$ has a node set formed by the landmark points in $\mathfrak{L}$, and a subset $\sigma$ of $\mathfrak{L}$ is in $\mathcal{W}(\mathfrak{L}, \mathcal{G})$ if and only if there exists a corresponding weak witness in the graph $\mathcal{G}$.
\end{definition}
\subsection{Wit-TopoPool: The Proposed Model}
\begin{figure*}[t!]
    \centering
    \includegraphics[width=0.9\textwidth]{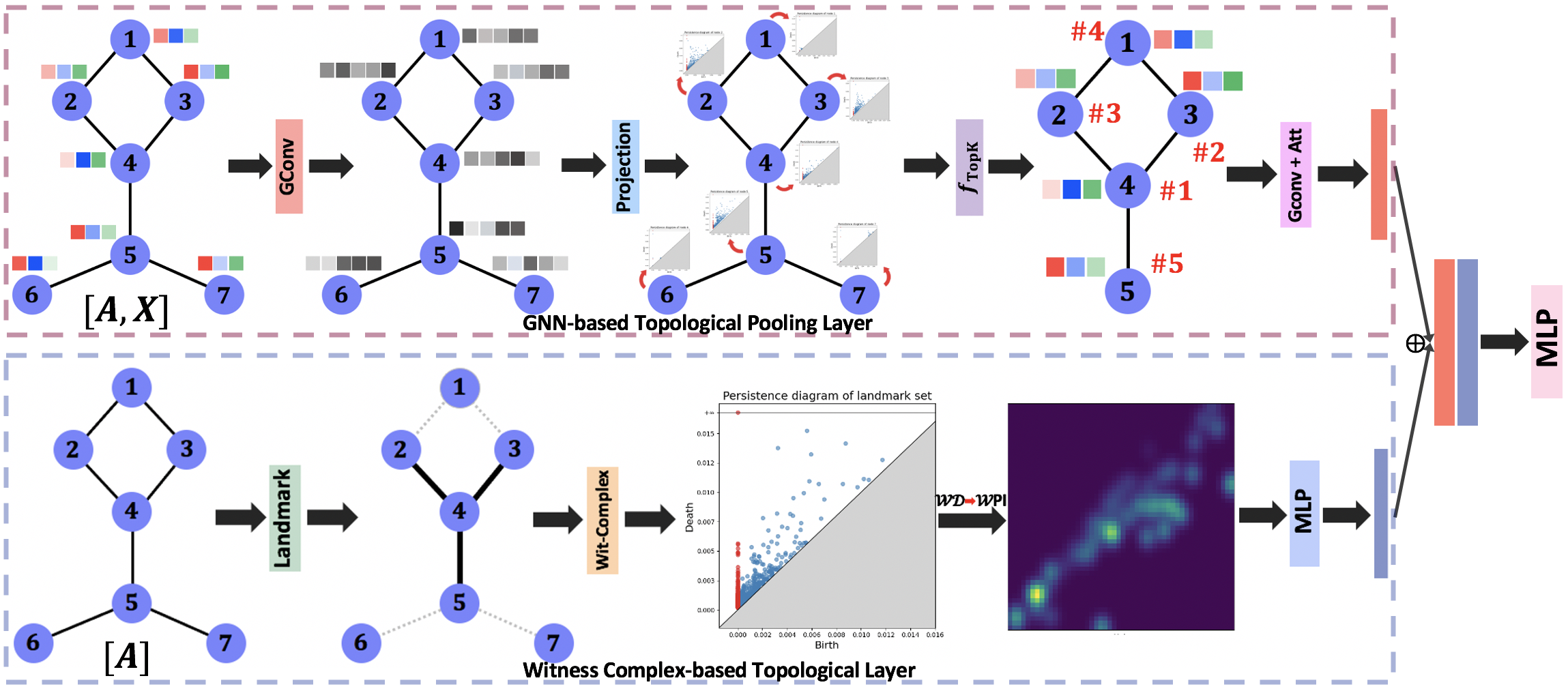}
    \caption{The overall architecture of Wit-TopoPool  (for more details see Appendix~A).\label{flowchart}}
\end{figure*}

We now introduce our Wit-TopoPool model which leverages two new topological concepts into graph learning, topological pooling and witness complex-based topological knowledge representation. The core idea of Wit-TopoPool is to simultaneously capture the discriminating topological information of the attributed graph $\mathcal{G}$, including its node feature representation, at both local and global levels. The first module of topological pooling assigns each node a topological score, based on measuring how valuable shape information of its local neighborhood is. In turn, the second module 
of witness complex-based topological knowledge representation allows us to learn the inherent global shape information of $\mathcal{G}$, by focusing only on the most essential landmarks of $\mathcal{G}$. The Wit-TopoPool architecture is illustrated in Figure~\ref{flowchart}.

\subsubsection{GNN-based Topological Pooling Layer}
The ultimate goal of this module is to sort nodes based on the importance of topological information, exhibited by their local neighborhoods at different learning stages (i.e., different layers of GNN). That is, a node neighborhood can be defined either based on the connectivity of the observed graph $\mathcal{G}$ or based on the similarity among node embeddings. This allows us to adaptively learn hidden higher-order structural dependencies among nodes that originally may not be within the proximity to each other with respect to the graph distance. 
To extract such local topological node representation, we first learn 
a latent node embedding by aggregating information from its the $k$-hop neighbors ($k \geq 1$) through graph convolutions
\begin{align}
\label{GNN_eq}
    {\boldsymbol{H}^{(\ell +1)}} = \sigma((\boldsymbol{\Tilde{D}}^{-\frac{1}{2}}\boldsymbol{\Tilde{A}}\boldsymbol{\Tilde{D}}^{\frac{1}{2}})^k\boldsymbol{H}^{(\ell)}\boldsymbol{W}^{(\ell)}).
\end{align}
Here $\sigma(\cdot)$ is the non-linear activation function (e.g., $\text{ReLU}(0, x) = \max{(x)}$), $\boldsymbol{W}^{(\ell)} \in \mathbb{R}^{d^\prime_c\times d_c}$ is trainable weight of $(\ell)$-th layer (where $d^\prime_c$ is the dimension of the $(\ell-1)$-th layer's output), $\boldsymbol{H}^{(0)} = \boldsymbol{X}$, $\boldsymbol{H}^{(\ell+1)} \in \mathbb{R}^{N \times d_c}$, and $k$-th power operator contains statistics from the $k$-th step of a random walk on the graph. 
Equipped with above node embedding, in order to capture the underlying structural information of nodes in latent feature space, we can measure the similarity between nodes. More specifically, for each node $u$, following Definition~(\ref{k_hop_neighbors_node_embedding}), we obtain the $\phi$-distance neighborhood subgraph $\boldsymbol{Z}^\phi_{u}$.
\begin{definition}[{\footnotesize $\phi$-distance Neighborhood of Node Embedding}]
\label{k_hop_neighbors_node_embedding}
\label{def3}
Let $\boldsymbol{H}^{(\ell+1)}$ be the node embedding of ($\ell$)-th layer of GNN. For any $u, v \in \mathcal{V}$, we can calculate the similarity score $\boldsymbol{Z}_{uv}$ between nodes $u$ and $v$ as (i) Cosine Similarity: $\boldsymbol{Z}_{uv} = \frac{\boldsymbol{H}_u^{(\ell+1)} \cdot \boldsymbol{H}_v^{(\ell+1)}}{||\boldsymbol{H}_u^{(\ell+1)}|| ||\boldsymbol{H}_v^{(\ell+1)}||}$ or (ii) Gaussian Kernel: $\boldsymbol{Z}_{uv} = \exp{(-\gamma||\boldsymbol{H}_u^{(\ell+1)} - \boldsymbol{H}_v^{(\ell+1)}||^2)}$ (where $\gamma$ is a free parameter). Given the pre-defined threshold $\phi > 0$, we have a $\phi$-distance neighborhood subgraph $\boldsymbol{Z}^\phi_u = (\mathcal{{V}}^\phi_u, \mathcal{{E}}^\phi_u)$, i.e., for any $v \in \mathcal{{V}}^\phi_u$, $\boldsymbol{Z}_{uv} \geq \phi$ and for any $v, w \in \mathcal{{V}}^\phi_u$ with $e_{vw} \in \mathcal{{E}}^\phi_u$, the similarity score between nodes $v$ and $w$ is larger than or equal to $\phi$, i.e., $\boldsymbol{Z}_{vw} \geq \phi$.
\end{definition}
Considering the $\phi$-distance neighborhood subgraph $\boldsymbol{Z}^\phi_{u}$ allows us to capture some hidden similarity between  unconnected yet relevant nodes, which in turn plays an important role for revealing high-order structural characteristics of the graph.
The intuition behind this idea is the following. Suppose that there are two people on a social media network whose graph distance in-between is high, but these people share some joint interests (node attributes). Then, while these two people are not neighbors in terms of Definition~(\ref{def1}), they might end up as neighbors in terms of Definition~(\ref{def3}).

We now derive a topological score of each node $u$ in terms of the importance of the topological information yielded by its new surrounding neighborhood. For each $u\in \mathcal{V}$, we first obtain its PD $\mathcal{D}_u=\text{Dg}(\mathcal{VR}(\boldsymbol{Z}^{\phi}_{u}))$. Then given the intuition that the longer the persistence of $\rho$ is, the more important the feature $\rho$ is, the {\it topological score} of node $u$ is defined via the persistence of its topological features in $\mathcal{D}_u$
\begin{align}
\label{weighting_function}
    y_{u} =
    \begin{cases}
    \sum_{\rho \in \mathcal{D}_u}(d_\rho - b_\rho) & (i)\\
    \sum_{\rho \in \mathcal{D}_u}\arctan{(C\times ((d_\rho - b_\rho)^\eta)} & (ii)
    \end{cases}.
\end{align}
Here (i) is an unweighted function and (ii) is a piecewise-linear weighted function, $C$ is a non-negative parameter, $\eta \geq 1$ defines a polynomial function, and $\arctan{(\cdot)}$ is a bounded and continuous function which is vital to guarantee the stability of the vectorization of the persistence diagram. (In Section~``Sensitivity Analysis'', we discuss experiments assessing  the impact of different topological score functions on graph classification tasks.) Finally, we sort all nodes in terms of their topological scores
$\boldsymbol{y} = [y_1, y_2, \dots, y_N]$ and select the top-$\ceil{\tau N}$ (where $\ceil{\cdot}$ is the operation of rounding up) nodes
\begin{align}
    \text{idx} = f_{\text{TopK}}(\boldsymbol{y}, \ceil{\tau N}),
\end{align}
where $f_{\text{TopK}} (\cdot)$ is a sorting function and produces indexes of $k$ nodes with largest topological scores. By taking the indices of above $\ceil{\tau N}$ nodes, the coarsened (pooled) adjacency matrix $\boldsymbol{A}_{\text{pool}} \in \mathbb{R}^{\ceil{\tau N} \times \ceil{\tau N}}$, indexed feature matrix $\boldsymbol{X}_{\text{pool}} \in \mathbb{R}^{\ceil{\tau N} \times F}$, and the topological pooling enhanced graph convolutional layer (TPGCL) can be formulated as
\begin{align}
    \boldsymbol{A}_{\text{pool}} & = \boldsymbol{A}[\text{idx}, \text{idx}], \boldsymbol{X}_{\text{pool}} = \boldsymbol{X}[\text{idx}, \text{idx}],\\
    \boldsymbol{H}_r & = \sigma((\boldsymbol{\Tilde{D}^{-\frac{1}{2}}_{\text{pool}}}\boldsymbol{\Tilde{A}_{\text{pool}}}\boldsymbol{\Tilde{D}^{\frac{1}{2}}_{\text{pool}}})\boldsymbol{X}_{\text{pool}}\boldsymbol{W}_{\text{pool}}),
\end{align}
where $\boldsymbol{\Tilde{A}}_{\text{pool}} = \boldsymbol{A}_{\text{pool}} + \boldsymbol{I}$ is the adjacency matrix with self-loop added to each node, $\boldsymbol{\Tilde{D}}_{\text{pool}}$ a diagonal matrix where $\boldsymbol{\Tilde{D}}_{\text{pool},uu} = \sum_{v}\boldsymbol{\Tilde{A}}_{\text{pool}, uv}$, $\boldsymbol{W}_{\text{pool}} \in \mathbb{R}^{F \times d_{\text{pool}}}$ is the learnable weight matrix in the topological pooling enhanced graph convolution operation (where $d_{\text{pool}}$ is the output dimension), and $\boldsymbol{H}_r \in \mathbb{R}^{\ceil{\tau N} \times d_{\text{pool}}}$ is the output of TPGCL.

Furthermore, to capture second-order statistics of pooled features and generate a global representation, we apply the attention mechanism (i.e., the second-order attention mechanism of~\citet{girdhar2017attentional}) on the TPGCL embedding $\boldsymbol{H}_r$ as follows
\begin{align}
\label{TP_output}
\boldsymbol{\hat{H}}_r = \boldsymbol{H}^{\top}_r (\boldsymbol{H}_r \boldsymbol{W}_r),
\end{align}
where $\boldsymbol{W}_r \in \mathbb{R}^{d_{\text{pool}} \times 1}$ is a trainable weight matrix and $\boldsymbol{\hat{H}}_r \in \mathbb{R}^{d_{\text{pool}}}$ is the final embedding.

\subsubsection{Witness Complex-based Topological Layer}
Now we turn to learning global shape characteristics of graph $\mathcal{G}$. To enhance computational efficiency and to focus on the most salient topological characteristics of $\mathcal{G}$, thereby mitigating the impact of noisy observations,
we propose a global topological representation learning module based on a witness complex $\mathcal{W}$ on a set of landmarks $\mathfrak{L}$.
Here we consider the landmark set $\mathfrak{L}$ obtained in one of three ways, i.e., (i) randomly, (ii) node degree centrality, and (iii) node betweenness centrality. Specifically, given a graph $\mathcal{G}$ with the number of nodes $N$ and some user-selected parameter $\psi$ ($\psi \in (0,1]$), the $\psi N$ landmarks can be selected (i) uniformly at random resulting in $\mathfrak{L}_r$; (ii) in the decreasing order of their degree centrality resulting in $\mathfrak{L}_d$;
and (iii) in the decreasing order of their betweenness centrality resulting in $\mathfrak{L}_b$.
Our goal is to select the most representative landmarks which can enhance the quality of the approximate simplicial representation. 
To adaptively learn the global topological information and correlations between topological structures and node features, (i) we first calculate the similarity matrix $\boldsymbol{S} \in \mathbb{R}^{N \times N}$ among $N$ nodes based on the node embedding of ($\ell$)-th GNN layer $\boldsymbol{H}^{(\ell+1)}$ (see Eq.~\ref{GNN_eq}) by using either cosine similarity (i.e., $\boldsymbol{S}_{uv} = \frac{\boldsymbol{H}_u^{(\ell+1)} \cdot \boldsymbol{H}_v^{(\ell+1)}}{||\boldsymbol{H}_u^{(\ell+1)}|| ||\boldsymbol{H}_v^{(\ell+1)}||}$) or Gaussian kernel (i.e., $\boldsymbol{S}_{uv} = \exp{(-\gamma||\boldsymbol{H}_u^{(\ell+1)} - \boldsymbol{H}_v^{(\ell+1)}||^2)}$), and then we preserve the connections between top similar pairs of nodes (e.g., $\boldsymbol{S}_{uv} \leq \zeta$, where $\zeta \geq 0$) and hence obtain a new graph structure $\mathcal{\hat{G}} = (\mathcal{V}, \mathcal{\hat{E}})$ (where $\mathcal{\hat{E}}$ depends on similarity matrix $\boldsymbol{S}$ and threshold $\zeta$); (ii) then armed with $\mathcal{\hat{G}}$, we extract persistent topological features and summarize them as persistence diagram $\mathcal{WD}_{\mathcal{\hat{G}}}$ of $\mathcal{\hat{G}}$, i.e., $\mathcal{WD}_{\mathcal{\hat{G}}} = \text{Dg}(\mathcal{W}(\mathfrak{L}, \mathcal{\hat{G}}))$.

To input the global topological information summarized by $\mathcal{WD}_{\mathcal{\hat{G}}}$ into neural network architecture, we convert $\mathcal{WD}_{\mathcal{\hat{G}}}$ to its finite-dimensional vector representation, i.e., witness complex persistence image of resolution $p$, i.e., $\mathcal{W}\text{PI}_{\mathcal{\hat{G}}}\in \mathbb{R}^{p \times p}$ (see Appendix~A for more details of $\mathcal{W}\text{PI}$) and then feed the $\mathcal{W}\text{PI}_{\mathcal{\hat{G}}}$ into multi-layer perceptron (MLP) to learn global topological information for graph embedding 
\begin{align}
\label{witness_complex_output}
    \boldsymbol{H}_w = \text{MLP}(\text{Flatten}(\mathcal{W}\text{PI}_\mathcal{\hat{G}}))
\end{align}
where $\text{Flatten}(\cdot)$ flattens 
$\mathcal{W}\text{PI}_{\mathcal{\hat{G}}}$ into an $p^2$-dimensional vector representation and $\boldsymbol{H}_w \in \mathbb{R}^{d_w}$ is the output of witness complex-based topological layer. Finally, we concatenate the outputs of GNN-based topological pooling layer (see Eq.~\ref{TP_output}) and witness complex-based topological layer (see Eq.~\ref{witness_complex_output}), and feed the concatenated vector into a single-layer MLP for classification as
\begin{align*}
    \boldsymbol{H}_o = \text{MLP}([\boldsymbol{\hat{H}}_r, \boldsymbol{H}_w]),
\end{align*}
where $[\cdot,\cdot]$ denotes the concatenation of the outputs of two layers and $\boldsymbol{H}_o$ is the final classification score.

\begin{table*}[h!]
\centering
\resizebox{2.\columnwidth}{!}{
\begin{tabular}{lcccccccc}
\toprule
\textbf{{Model}} &\textbf{{BZR}} & \textbf{{COX2}} & \textbf{{MUTAG}} & \textbf{{PROTEINS}} &\textbf{{PTC\_MR}} & \textbf{PTC\_MM} & \textbf{PTC\_FM} & \textbf{PTC\_FR}\\
\midrule
CSM~\cite{kriege2012subgraph} &84.54$\pm$0.65 &79.78$\pm$1.04 &87.29$\pm$1.25 &OOT& 58.24$\pm$2.44 &63.30$\pm$1.70 & 63.80$\pm$1.00 & 65.51$\pm$9.82\\
HGK-SP~\cite{morris2016faster}&81.99$\pm$0.30 &78.16$\pm$0.00& 80.90$\pm$0.48 &74.53$\pm$0.35& 57.26$\pm$1.41 & 57.52$\pm$9.98 & 52.41$\pm$1.79 & 66.91$\pm$1.46\\
HGK-WL~\cite{morris2016faster}& 81.42$\pm$0.60 &78.16$\pm$0.00 &75.51$\pm$1.34 & 74.53$\pm$0.35& 59.90$\pm$4.30 & 67.22$\pm$5.98 & 64.72$\pm$1.66 &  67.90$\pm$1.81\\
WL~\cite{shervashidze2011weisfeiler}& 86.16$\pm$0.97& 79.67$\pm$1.32& 85.75$\pm$1.96 & 73.06$\pm$0.47& 57.97$\pm$0.49 & 67.28$\pm$0.97 & 64.80$\pm$0.85& 67.64$\pm$0.74\\
WL-OA~\cite{kriege2016valid} & \dotuline{87.43$\pm$0.81} &81.08$\pm$0.89 & 86.10$\pm$1.95 & 73.50$\pm$0.87&62.70$\pm$1.40 & 66.60$\pm$1.16 & 66.28$\pm$1.83 & 67.82$\pm$5.03\\
DGCNN~\cite{zhang2018end} & 79.40$\pm$1.71 & 79.85$\pm$2.64 & 85.83$\pm$1.66 & 75.54$\pm$0.94 &  58.59$\pm$2.47 & 62.10$\pm$14.09 & 60.28$\pm$6.67 & 65.43$\pm$11.30\\
GCN~\cite{kipf2016semi} & 79.34$\pm$2.43& 76.53$\pm$1.82 & 80.42$\pm$2.07 & 70.31$\pm$1.93&62.26$\pm$4.80 & 67.80$\pm$4.00 & 62.39$\pm$0.85 & 69.80$\pm$4.40\\
GIN~\cite{xu2018powerful} & 85.60$\pm$2.00 & 80.30$\pm$5.17 &  89.39$\pm$5.60 & 76.16$\pm$2.76 & 64.60$\pm$7.00 & 67.18$\pm$7.35 & 64.19$\pm$2.43 & 66.97$\pm$6.17\\
Top-$K$~\cite{gao2019graph} & 79.40$\pm$1.20 & 80.30$\pm$4.21 & 67.61$\pm$3.36 & 69.60$\pm$3.50 & 64.70$\pm$6.80 & 67.51$\pm$5.96 & 65.88$\pm$4.26 & 66.28$\pm$3.71 \\
MinCutPool~\cite{bianchi2020spectral} &  82.64$\pm$5.05 & 80.07$\pm$3.85 & 79.17$\pm$1.64 & 76.52$\pm$2.58 & 64.16$\pm$3.47 & N/A & N/A  & N/A  \\
DiffPool~\cite{ying2018hierarchical} & 83.93$\pm$4.41 & 79.66$\pm$2.64 & 79.22$\pm$1.02 & 73.63$\pm$3.60 & 64.85$\pm$4.30 & 66.00$\pm$5.36 & 63.00$\pm$3.40 & \dotuline{69.80$\pm$4.40}\\
EigenGCN~\cite{ma2019graph} & 83.05$\pm$6.00 & 80.16$\pm$5.80 & 79.50$\pm$0.66 & 74.10$\pm$3.10 & N/A  & N/A  & N/A & N/A  \\
SAGPool~\cite{lee2019self} & 82.95$\pm$4.91 & 79.45$\pm$2.98 & 76.78$\pm$2.12 & 71.86$\pm$0.97 &69.41$\pm$4.40  &66.67$\pm$8.57 & 67.65$\pm$3.72 & 65.71$\pm$10.69\\
HaarPool~\cite{wang2020haar} & 83.95$\pm$5.68 & \dotuline{82.61$\pm$2.69} & \dotuline{90.00$\pm$3.60} &73.23$\pm$2.51 & 66.68$\pm$3.22 &69.69$\pm$5.10 & 65.59$\pm$5.00& 69.40$\pm$5.21 \\
PersLay~\cite{carriere2020perslay} & 82.16$\pm$3.18 &80.90$\pm$1.00 &89.80$\pm$0.90 & 74.80$\pm$0.30 & N/A & N/A  & N/A  & N/A \\
FC-V~\cite{o2021filtration}& 85.61$\pm$0.59&81.01$\pm$0.88 & 87.31$\pm$0.66&74.54$\pm$0.48& N/A  & N/A & N/A  & N/A \\
MPR~\cite{bodnar2021deep} & N/A  & N/A & 84.00$\pm$8.60 & 75.20$\pm$2.20 & 66.36$\pm$6.55 & 68.60$\pm$6.30 & 63.94$\pm$5.19& 64.27$\pm$3.78 \\
SIN~\cite{bodnar2021weisfeiler} & N/A  & N/A  & N/A  & \dotuline{76.50$\pm$3.40} & \dotuline{66.80$\pm$4.56} & \dotuline{70.55$\pm$4.79} & \dotuline{68.68$\pm$6.80} & 69.80$\pm$4.36 \\
\midrule
\textbf{Wit-TopoPool (ours)} &{\bf 87.80$\pm$2.44} &\hspace{-2ex}$^{***}${\bf 87.24$\pm$3.15} & {\bf 93.16$\pm$4.11} &$^{**}${\bf 80.00$\pm$3.22} &$^{*}${\bf 70.57$\pm$4.43} & \hspace{-2ex}$^{***}${\bf 79.12$\pm$4.45} & {\bf 71.71$\pm$4.86} & \hspace{-2ex}$^{***}${\bf 75.00$\pm$3.51}\\
\bottomrule
\end{tabular}}
\caption{Performance on molecular and chemical graphs. The best results are given in {\bf bold} while the best performances achieved by the runner-ups are \dotuline{underlined}. 
\label{classification_results_bio_graphs}}
\end{table*}

\section{Experiments}
\paragraph{Datasets} We validate Wit-TopoPool on graph classification tasks using the following 11 real-world graph datasets (for further
details, please refer to Appendix B): (i) 3 chemical compound datasets: MUTAG, BZR, and COX2, where graphs represent chemical compounds, nodes are different atoms, and edges are chemical bonds; (ii) 5 molecular compound datasets: PROTEINS, PTC\_MR, PTC\_MM, PTC\_FM, and PTC\_FR, where nodes are secondary structure elements and edge existence between two nodes implies that the nodes are adjacent nodes in an amino acid sequence or three nearest-neighbor interactions; (iii) 2 internet movie databases: IMDB-BINARY (IMDB-B) and IMDB-MULTI (IMDB-M), where nodes are actors/actresses and there is an edge if the two people appear in the same movie, and (iv) 1 Reddit (an online aggregation and discussion website) discussion threads dataset: REDDIT-BINARY (REDDIT-B), where nodes are Reddit users and edges are direct replies in the discussion threads. Each dataset includes multiple graphs of each class, and we aim to classify graph classes. For all graphs, we use different random seeds for 90/10 random training/test split. Furthermore, we perform a one-sided two-sample $t$-test between the best result and the best performance achieved by the runner-up, where *, **, *** denote significant, statistically significant, highly statistically significant results, respectively.

\paragraph{Baselines} We evaluate the performances of our Wit-TopoPool on 11 graph datasets versus 18 state-of-the-art baselines (including 4 types of approaches): (i) 6 graph kernel-based methods: (1) comprised of the subgraph matching kernel (CSM)~\cite{kriege2012subgraph}, (2) Shortest
Path Hash Graph Kernel (HGK-SP)~\cite{morris2016faster}, (3) Weisfeiler–Lehman Hash
Graph Kernel (HGK-WL)~\cite{morris2016faster}, (4) Weisfeiler–Lehman (WL)~\cite{shervashidze2011weisfeiler}, and (5) Weisfeiler-Lehman Optimal Assignment (WL-OA)~\cite{kriege2016valid}; (ii) 3 GNNs: (6) Graph Convolutional Network (GCN)~\cite{kipf2016semi}, (7) Graph Isomorphism Network (GIN)~\cite{xu2018powerful}, and (8) Deep Graph Convolutional Neural Network (DGCNN)~\cite{zhang2018end}; (iii) 4 topological and simplicial complex-based methods: (9) Neural Networks for Persistence Diagrams (PersLay)~\cite{carriere2020perslay}, (10) Filtration Curves with a Random Forest (FC-V)~\cite{o2021filtration}, (11) Deep Graph Mapper (MPR)~\cite{bodnar2021deep}, and (12) Message Passing Simplicial Networks (SIN)~\cite{bodnar2021weisfeiler}; (iv) 6 graph pooling methods: (13) GNNs with Differentiable Pooling (DiffPool)~\cite{ying2018hierarchical}, (14) TopKPooling with Graph U-Nets (Top-$K$)~\cite{gao2019graph}, (15) GCNs with Eigen Pooling (EigenGCN)~\cite{ma2019graph}, (16) Self-attention Graph Pooling (SAGPool)~\cite{lee2019self}, (17) Spectral Clustering for Graph Pooling (MinCutPool)~\cite{bianchi2020spectral}, and (18) Haar Graph Pooling (HaarPool)~\cite{wang2020haar}.

\paragraph{Experiment Settings}
We conduct our experiments on two NVIDIA GeForce RTX 3090 GPU cards with 24GB memory. Wit-TopoPool is trained end-to-end by using Adam optimizer and the optimal trainable weight matrices are trained by minimizing the cross-entropy loss function. The tuning of Wit-TopoPool on each dataset is done via grid hyperparameter configuration search over a fixed set of choices and the same cross-validation setup is used to tune baselines. In our experiments, for all datasets, we set the grid size of $\mathcal{W}$PI to $5 \times 5$, and the MLP is of 2 layers where
Batchnorm and Dropout with dropout ratio of $p_{drop} \in \{0, 0.1, \dots, 0.5\}$ applied after the fist layer of MLP. For MUTAG, the number of layers in the neural networks and hidden feature dimension is set to be 3 and 64 respectively. For BZR, the number of layers in the neural networks and hidden feature dimension is set to be 5 and 16 respectively. For COX2 and IMDB-M, the number of layers in the neural networks and hidden feature dimension is set to be 3 and 8 respectively. For PROTEINS and PTC\_MR, the number of layers in the neural networks and hidden feature dimension is set to be 5 and 8 respectively. For PTC\_MM, PTC\_FM, PTC\_FR, IMDB-B, and REDDIT-B, the number of layers in the neural networks and hidden feature dimension is set to be 5 and 32 respectively. The grid search spaces for learning rate and hidden size of $\mathcal{W}\text{PI}_{\mathcal{\hat{G}}}$ representation learning are $l_{r} \in \{0.001, 0.003, 0.005, 0.008, 0.01, 0.05\}$ and $d_w \in \{8, 16, 32, 64, 128\}$, respectively. The range of grid search space for the hyperparameter $\psi$ (for the number of landmark points) is searched in $\{0.1, 0.2, 0.3, 0.5, 0.6\}$. For BZR and COX2, the batch sizes are 64 and 16, respectively; for other graph datasets, we train our network with batch size 8. Parameter $C$ in Eq.~\ref{weighting_function} {\it (ii)} is chosen from values $\{0.1, 0.2, \dots, 0.5\}$ and we set $\eta$ in Eq.~\ref{weighting_function} {\it (ii)} to 2. The source code is available at~\url{https://github.com/topologicalpooling/TopologicalPool.git}.

\subsection{Experiment Results}
The evaluation results on 11 graph datasets are summarized in Tables~\ref{classification_results_bio_graphs} and~\ref{classification_results_social_graphs}.  We also conduct ablation studies to assess contributions of the key Wit-TopoPool components. Moreover, we perform sensitivity analysis to examine the impact of different choices for topological score functions and landmark sets. OOM indicates out of memory (from an allocation of 128 GB RAM) and OOT indicates out of time (within 120 hours).

\paragraph{Molecular and Chemical Graphs} Table~\ref{classification_results_bio_graphs} shows the performance comparison among 18 baselines on BZR, COX2, MUTAG, PROTEINS, and four PTC datasets with different carcinogenicities on rodents (i.e., PTC\_MR, PTC\_MM, PTC\_FM, and PTC\_FR) for graph classification. Our Wit-TopoPool consistently outperforms baseline models on all 8 datasets. In particular, the average relative gain of Wit-TopoPool over the runner-ups is 5.10\%. The results demonstrate the effectiveness of Wit-TopoPool. In terms of baseline models, graph kernels only account for the graph structure information and tend to suffer from higher computational costs. In turn, GNN-based models, e.g., GIN, capture both local graph structures and information of the neighborhood for each node, hence, resulting in improvement over graph kernels. Comparing with GNN-based models, graph pooling methods such as SAGPool and HaarPool utilize the hierarchical structure of the graph and extract important geometric information on the observed graphs. Finally, PersLay, FC-V, MPR, and SIN are the state-of-the-art topological and simplicial complex-based models, specialized on extracting topological information and higher-order structures from the observed graphs. A common limitation of these approaches is that they do not simultaneously capture both local and global topological properties of the graph. Hence, it is not surprising that performance of Wit-TopoPool which systematically integrates all types of the above information on the observed graphs is substantially higher than that of the benchmark models.

\paragraph{Social Graphs} Table~\ref{classification_results_social_graphs} shows the performance comparison on 3 social graph datasets. Similarly, Table~\ref{classification_results_social_graphs} indicates that our Wit-TopoPool model is always better than baselines for all social graph datasets. We find that, even compared to the baselines (which feeds neural networks with topological summaries (i.e., PersLay) or integrates higher-order structures into GNNs (i.e., SIN)), Wit-TopoPool is highly competitive, revealing that global and local topological representation learning modules can enhance the model expressiveness.

\begin{table}[h!]
\centering
\resizebox{1.\columnwidth}{!}{
\begin{tabular}{lccc}
\toprule
\textbf{{Model}} & \textbf{{IMDB-B}}&\textbf{{IMDB-M}}&\textbf{{REDDIT-B}} \\
\midrule
CSM~\cite{kriege2012subgraph} &OOT&OOT&OOT\\
HGK-SP~\cite{morris2016faster} & 73.34$\pm$0.47&51.58$\pm$0.42 & OOM \\
HGK-WL~\cite{morris2016faster}& 72.75$\pm$1.02&50.73$\pm$0.63 & OOM \\
WL~\cite{shervashidze2011weisfeiler}& 71.15$\pm$0.47 & 50.25$\pm$0.72 & 77.95$\pm$0.60\\
WL-OA~\cite{kriege2016valid} &74.01$\pm$0.66 & 49.95$\pm$0.46 &87.60$\pm$0.33 \\
DGCNN~\cite{zhang2018end} &70.00$\pm$0.90 &47.80$\pm$0.90 & 76.00$\pm$1.70 \\
GCN~\cite{kipf2016semi} &66.53$\pm$2.33 & 48.93$\pm$0.88 & 89.90$\pm$1.90\\
GIN~\cite{xu2018powerful} & 75.10$\pm$5.10 & 52.30$\pm$2.80 & 92.40$\pm$2.50\\
Top-$K$~\cite{gao2019graph} &73.17$\pm$4.84 &48.80$\pm$3.19 & 79.40$\pm$7.40\\
MinCutPool~\cite{bianchi2020spectral} &70.77$\pm$4.89 & 49.00$\pm$2.83 & 87.20$\pm$5.00\\
DiffPool~\cite{ying2018hierarchical} & 68.60$\pm$3.10 & 45.70$\pm$3.40 & 79.00$\pm$1.10\\
EigenGCN~\cite{ma2019graph} &70.40$\pm$3.30 & 47.20$\pm$3.00 & N/A \\
SAGPool~\cite{lee2019self} &74.87$\pm$4.09 &49.33$\pm$4.90 & 84.70$\pm$4.40\\
HaarPool~\cite{wang2020haar} &73.29$\pm$3.40  &49.98$\pm$5.70  & N/A  \\
PersLay~\cite{carriere2020perslay} &71.20$\pm$0.70 & 48.80$\pm$0.60 & N/A \\
FC-V~\cite{o2021filtration}& 73.84$\pm$0.36 & 46.80$\pm$0.37 & 89.41$\pm$0.24\\
MPR~\cite{bodnar2021deep} &73.80$\pm$4.50 & 50.90$\pm$2.50 &  86.20$\pm$6.80\\
SIN~\cite{bodnar2021weisfeiler} & \dotuline{75.60$\pm$3.20} & \dotuline{52.50$\pm$3.00} & \dotuline{92.20$\pm$1.00} \\
\midrule
\textbf{Wit-TopoPool (ours)} & \hspace{-3ex}$^{***}${\bf 78.40$\pm$1.50} & {\bf 53.33$\pm$2.47} & \hspace{-1ex}$^{*}${\bf 92.82$\pm$1.10}\\
\bottomrule
\end{tabular}}
\caption{Performance on social graphs. The best results are given in {\bf bold} while the best performances achieved by the runner-ups are \dotuline{underlined}. 
\label{classification_results_social_graphs}}
\end{table}

\subsection{Ablation Study}
To evaluate the contributions of the different components in our Wit-TopoPool model, we perform exhaustive ablation studies on COX2, PTC\_MM, and IMDB-B datasets. We use Wit-TopoPool as the baseline architecture and consider three ablated variants: (i) Wit-TopoPool without topological pooling enhanced graph convolutional layer (W/o TPGCL), (ii) Wit-TopoPool without witness complex-based topological layer (W/o Wit-TL), and (iii) Wit-TopoPool without attention mechanism (W/o Attention Mechanism). The experimental results are shown in Table~\ref{ablation_architecture} and we prove the validity of each component. As  Table~\ref{ablation_architecture} suggest, we find that (i) ablating each of above component leads to the performance drops in comparison with the full Wit-TopoPool model, thereby, indicating that each of the designed components contributes to the success of Wit-TopoPool, (ii) on all three datasets, TPGCL module significantly improves the classification results, i.e., Wit-TopoPool outperforms Wit-TopoPool w/o TPGCL with an average relative gain 5.30\% over three datasets -- this phenomenon implies that learning both local topological information and node features are critical for successful graph learning, (iii) in comparison to Wit-TopoPool and Wit-TopoPool w/o Wit-TL, Wit-TopoPool always outperforms because the Wit-TL module enables the model to effectively incorporate more global topological information, demonstrating the significance of the proposed global topological representation learning module for graph classification, and (iv) Wit-TopoPool consistently outperforms Wit-TopoPool w/o Attention Mechanism on all 3 datasets, indicating the attention mechanism can successfully extract the most correlated information and, hence, improves the generalization of unseen graph structures. Moreover, we also compare Wit-TopoPool with VR-TopoPool (i.e., replacing witness complex in global information learning with Vietoris-Rips complex) (see Appendix~B for a discussion).

\begin{table}[h!]
\centering
\setlength\tabcolsep{1.pt}
\resizebox{1.\columnwidth}{!}{
\begin{tabular}{llc}
\toprule
&\textbf{Architecture} & \textbf{Accuracy mean$\pm$std} \\
\midrule
\multirow{4}{*}{\textbf{COX2}}&\textbf{Wit-TopoPool} &\hspace{-1ex}$^{*}${\bf 87.24$\pm$3.15} \\
&W/o TPGCL & 82.67$\pm$3.26\\
&W/o Wit-TL & 85.21$\pm$3.20\\
&W/o Attention mechanism & \dotuline{85.58$\pm$3.53}\\
\midrule
\multirow{4}{*}{\textbf{PTC\_MM}}&\textbf{Wit-TopoPool} & {\bf 76.76$\pm$5.78} \\
&W/o TPGCL & 67.38$\pm$5.33\\
&W/o Wit-TL & 70.58$\pm$5.29\\
&W/o Attention mechanism & \dotuline{75.12$\pm$5.59}\\
\midrule
\multirow{4}{*}{\textbf{IMDB-B}}&\textbf{Wit-TopoPool} & \hspace{-2ex}$^{**}${\bf 78.40$\pm$1.50} \\
&W/o TPGCL & 73.93$\pm$1.83\\
&W/o Wit-TL & \dotuline{77.00$\pm$1.69}\\
&W/o Attention mechanism & 76.20$\pm$1.98\\
\bottomrule
\end{tabular}}
\caption{Ablation study of the Wit-TopoPool architecture.\label{ablation_architecture}}
\end{table}

\paragraph{Sensitivity Analysis} We perform sensitivity analysis of (i) landmark set selection and (ii) topological score function to explore the effect of above two components on our Wit-TopoPool performance. The optimal choice of landmark set selection and topological score function can be obtained via cross-validation. We first explore the effect of landmark set selection. We consider 3 types of landmark set selections, i.e., (i) randomly ($\mathfrak{L}_r$), (ii) node betweenness centrality ($\mathfrak{L}_b$), and (iii) node degree centrality ($\mathfrak{L}_d$), and report results on COX2 and PTC\_MM datasets. As Table~\ref{landmark_set_sensitivity} shows, we observe that the landmark set selection based on either node betweenness or degree centrality helps to improve the graph classification performance, whereas the landmark set based on randomly results in the performance drop. We also explore the effect of topological score function for the importance measurement of the persistence diagram (see Eq.~\ref{weighting_function}). As the results in Table~\ref{importance_measurements} suggest, summing over lifespans of topological features (points) in persistence diagrams can significantly improve performance, but applying piecewise linear weighting function on topological features may result in deterioration of performance.  

\begin{table}[h!]
\centering
\begin{tabular}{lcc}
\toprule
\textbf{Dataset}&\textbf{\footnotesize Landmark set} & \textbf{\footnotesize Accuracy mean$\pm$std} \\
\midrule
\multirow{3}{*}{\textbf{\footnotesize COX2}}& $\mathfrak{L}_r$ &82.98$\pm$3.88\\
& $\mathfrak{L}_b$ & \dotuline{85.10$\pm$2.52}\\
& $\mathfrak{L}_d$ & {\bf 87.24$\pm$3.15}\\
\midrule
\multirow{3}{*}{\textbf{\footnotesize PTC\_MM}}& $\mathfrak{L}_r$ &71.53$\pm$6.17\\
& $\mathfrak{L}_b$ & {\bf 79.12$\pm$4.45}\\
& $\mathfrak{L}_d$ & \dotuline{76.76$\pm$5.78}\\
\bottomrule
\end{tabular}
\caption{Sensitivity analysis with respect to the landmark set selection for Wit-TopoPool on COX2 and PTC\_MM.\label{landmark_set_sensitivity}}
\end{table}

\begin{table}[h!]
\centering
\setlength\tabcolsep{3pt}
\begin{tabular}{lcc}
\toprule
\textbf{Dataset}&\textbf{\footnotesize Weighting function} & \textbf{\footnotesize Accuracy mean$\pm$std} \\
\midrule
\multirow{2}{*}{\textbf{\footnotesize COX2}}& {\small $d_\rho - b_\rho$} & \hspace{-3ex}$^{***}${\bf 87.24$\pm$3.15}\\
&  {\small $\arctan{(C\times ((d_\rho - b_\rho)^\eta)}$} & 79.78$\pm$1.06\\
\midrule
\multirow{2}{*}{\textbf{\footnotesize PTC\_MM}}& {\small $d_\rho - b_\rho$}  & \hspace{-1ex}$^{*}${\bf 79.12$\pm$4.45}\\
& {\small $\arctan{(C\times ((d_\rho - b_\rho)^\eta)}$} & 76.18$\pm$5.00\\
\bottomrule
\end{tabular}
\caption{Sensitivity analysis with respect to selection of weighting functions within the topological score for Wit-TopoPool on COX2 and PTC\_MM.\label{importance_measurements}}
\end{table}

\paragraph{Computational Complexity} Computational complexity of the standard persistent homology matrix reduction algorithm~\cite{edelsbrunner2000topological} (i.e., based on column operations over boundary matrix of the complex) runs in cubic time in the worst case, i.e., $\mathcal{O}(m^3)$, where $m$ is the number of simplices in the filtration. For 0-dimensional PH, it can be computed efficiently using disjoint sets with complexity $\mathcal{O}(m\alpha^{-1}m)$, where $\alpha^{-1}(\cdot)$ is the inverse Ackermann function~\cite{cormen2022introduction}. Computational complexity of the witness complex construction is $\mathcal{O}(\mathfrak{L} \log{(n)})$ (where $n$ is the number of data points and $\mathfrak{L}$ is the landmark set), involving calculating the distance between data points and landmark points.

\section{Conclusion}
In this paper, we have proposed Wit-TopoPool, a differentiable and comprehensive pooling operator for graph classification that simultaneously extracts the key topological characteristics of graphs at both local and global levels, using the notions of persistence, landmarks, and witnesses. In the future, we will expand the ideas of learnable topological representations and adaptive similarity learning among nodes to dynamic and multilayer networks.
\section{Acknowledgements}
This work was supported by the NSF grant \# ECCS 2039701 and ONR grant \# N00014-21-1-2530. Part of this material is also based upon work supported by (while serving at) the NSF. The views expressed in the article do not necessarily represent the views of NSF and ONR.

\bibliography{topopool}

\clearpage
\appendix
\section{A. Additional Details of Wit-TopoPool Architecture}
\subsection{A.1. Witness Complex-based Persistence Image}
Inspired by the notion of a persistent image as a stable summary of ordinary persistence~\cite{adams2017persistence}, we propose a representation of $\mathcal{WD}$ as {\it witness complex-based persistence image} ($\mathcal{W}\text{PI}$) in order to integrate topological features summarized by witness complex-based persistence diagram ($\mathcal{WD}$) into the designed topological layer. Formally, the process of $\mathcal{W}\text{PI}$ generation is formulated as follows
\begin{itemize}[leftmargin=*]
\item\textit{Step 1:} Map a witness complex-based persistence diagram $\mathcal{WD}$ to an integrable function $\Psi_{\mathcal{WD}}: \mathbb{R}^2 \rightarrow \mathbb{R}$, called a {\it witness complex-based persistence surface}. The witness complex-based persistence surface is given by sums of weighted probability density functions (here we consider Gaussian functions) that are centered at each point in $\mathcal{WD}$, and it is defined as
$$\Psi_{\mathcal{WD}} = \sum_{\mu \in T(\mathcal{WD})}f\left(\mu\right)e^{\bigl\{-\frac{{||z-\mu||^2}}{{2\xi^2}}\bigr\}},$$
where $T(\mathcal{WD})$ is the transformed multi-set in $\mathcal{WD}$, i.e., $T(\mathcal{WD} (x,y)) = (x,y-x)$; $f(\mu)$ is a non-negative weighting function with mean $\mu = (\mu_x, \mu_y) \in \mathbb{R}^2$ and variance $\xi^2$, which depends on the distance from the diagonal.

\item \textit{Step 2:} Take a discretization 
 of a subdomain of zigzag persistence surface $\Psi_{\mathcal{WD}}$ in a grid. Finally, the matrix of pixel values $\mathcal{W}\text{PI}$ can be obtained by subsequent integration on each grid box.
\end{itemize}

\subsection{A.2. The Wit-TopoPool Architecture}
The overall architecture of Wit-TopoPool is illustrated in Figure~\ref{flowchart}. Wit-TopoPool consists of 3 components: ({\bf i}) the top row shows the procedure of GNN-based topological pooling layer (TPGCL): TPGCL first generate node embedding $\boldsymbol{H}^{(\ell + 1)}$ through graph convolutions based on graph structure ($\boldsymbol{A}$) and node features ($\boldsymbol{X}$); after that, for each node, it generates $\phi$-distance neighborhood subgraph by calculating similarity relations between the target node and the rest of nodes (see Definition~3), and then extracts topological features (i.e., persistence diagram); in the next step, it selects top-$K$ nodes based on the importance of topological information and feeds the sorted top-$K$ node embedding to graph convolutional layer and attention mechanism; ({\bf ii}) the bottom row shows the procedure of witness complex-based topological layer (Wit-TL): based on graph structure, it first select the landmark set, and then applies witness complex over the landmark set to produce simplices -- this step enables us to obtain topological features (witness complex-based persistence diagram and witness complex-based persistence image); after that Wit-TL feeds the (flattened) witness complex-based persistence image to MLP and learns global topological information; ({\bf iii}) finally, the outputs of above two modules are concatenated and the concatenated output is fed into MLP for graph classification evaluation.

\section{B. Datasets and Additional Experiments}
\begin{table}[h!]
\centering
\resizebox{0.8\columnwidth}{!}{
\setlength\tabcolsep{4pt}
\begin{tabular}{lccccc}
\toprule
\textbf{{Dataset}} & \textbf{{\# Graphs}} &\textbf{{Avg.} $|\mathcal{V}|$} & \textbf{{Avg.} $|\mathcal{E}|$} & \textbf{{\# Class}} \\
\midrule
MUTAG &188 &17.93 &19.79 &2 \\
BZR &405 &35.75 &38.35 &2 \\
COX2 &467 &41.22 &43.45 &2 \\
PROTEINS &1113 &39.06 &72.82 &2 \\
PTC\_MR &  344 & 14.29 & 14.69 & 2\\
PTC\_MM &  336 & 13.97 & 14.32 & 2\\
PTC\_FM &  349 & 14.11 & 14.48 & 2\\
PTC\_FR & 351 & 14.56 & 15.00 & 2 \\
IMDB-B &1000 &19.77 &96.53 &2 \\
IMDB-M & 1500 & 13.00 & 65.94 & 3 \\
REDDIT-B & 2000 &  429.63 & 497.75& 2 \\
\bottomrule
\end{tabular}}
\caption{Summary statistics of the benchmark datasets.\label{tab:datasets}}
\end{table}

To better justify the efficiency and effect of witness complex in our proposed architecture, we conduct experiments on comparing between (i) integrating global topological information based on witness complex into the topological layer (i.e., Wit-TopoPool) and (ii) integrating global topological information based on Vietoris-Rips ($\mathcal{VR}$) complex into the topological layer (i.e., $\mathcal{VR}$-TopoPool) on COX2, PTC\_MM, and IMDB-B datasets. Table~\ref{ablation_witness_vr} reports the performances of Wit-TopoPool and $\mathcal{VR}$-TopoPool, and average running times (in seconds) of persistent homology (PH) computation and training time per epoch. As Table~\ref{ablation_witness_vr} shows, ({\bf i}) on COX2 and PTC\_MM, Wit-TopoPool and  $\mathcal{VR}$-TopoPool achieve similar performance; however, on IMDB-B dataset, Wit-TopoPool significantly outperforms $\mathcal{VR}$-TopoPool; moreover, Wit-TopoPool always yields lower standard deviation than $\mathcal{VR}$-TopoPool on all three datasets, which we conjecture is due to the reason that the landmark set selection helps to select/extract more representative nodes in coarser level and hence alleviates the impact of topological noise;
({\bf ii}) from a computational perspective, as expected, both the running time of witness complex and training time per epoch of Wit-TopoPool are shorter than $\mathcal{VR}$-complex and $\mathcal{VR}$-TopoPool, respectively -- due to the reason that witness complex approximates the $\mathcal{VR}$-TopoPool by constructing simplicial complexes over the landmark set $\mathfrak{L}$ (i.e., a subset of vertex set, $\mathfrak{L} \subseteq \mathcal{V}$).

\begin{table}[h!]
\centering
\setlength\tabcolsep{1.pt}
\resizebox{0.99\columnwidth}{!}{
\begin{tabular}{llccc}
\toprule
\multirow{2}{*}{\textbf{Dataset}} & \multirow{2}{*}{\textbf{Architecture}} &
\multirow{2}{*}{\textbf{Accuracy mean$\pm$std}} & \multicolumn{2}{c}{\textbf{Average time taken (sec)}} \\
& & & PH & Training time (epoch)\\
\midrule
\multirow{2}{*}{\textbf{COX2}}&{Wit-TopoPool} &  87.24$\pm$3.15 & $3.75 \times 10^{-3}$ & 11.09\\
&$\mathcal{VR}$-TopoPool &87.63$\pm$5.00 & $4.26 \times 10^{-3}$ & 12.72 \\
\midrule
\multirow{2}{*}{\textbf{PTC\_MM}}&{Wit-TopoPool} & 76.76$\pm$5.78 & $2.39 \times 10^{-3}$ & 3.60\\
&$\mathcal{VR}$-TopoPool &76.17$\pm$6.47 & $3.81 \times 10^{-3}$ & 3.98\\
\midrule
\multirow{2}{*}{\textbf{IMDB-B}}&{Wit-TopoPool} &  78.40$\pm$1.50 & $3.11 \times 10^{-3}$ &4.36 \\
&$\mathcal{VR}$-TopoPool &71.10$\pm$2.55 &$3.57 \times 10^{-3}$ & 4.86\\
\bottomrule
\end{tabular}
}
\caption{Comparisons of witness complex vs. $\mathcal{VR}$-complex. PH means persistent homology.\label{ablation_witness_vr}}
\end{table}
\end{document}